\documentclass[10pt,conference]{IEEEtran}

\usepackage{amsmath,amssymb}
\usepackage{algorithm}
\usepackage{algpseudocode}
\usepackage{graphicx}
\usepackage{multirow}
\usepackage{float}
\usepackage{rotating}
\usepackage{placeins}
\usepackage{hyperref}

\title{RCMAES: A Robust CMA-ES Variant for CEC2026 Competition}

\author{
	\IEEEauthorblockN{
		Khoirul Faiq Muzakka\IEEEauthorrefmark{1}\IEEEauthorrefmark{2},
		S\"oren M\"oller\IEEEauthorrefmark{1},
		Martin Finsterbusch\IEEEauthorrefmark{1}
	}
	\IEEEauthorblockA{\IEEEauthorrefmark{1}
		Institute of Energy Materials and Devices (IMD-2), Forschungszentrum J\"ulich GmbH, Germany
	}
	\IEEEauthorblockA{\IEEEauthorrefmark{2}
		Email: k.muzakka@fz-juelich.de
	}
}

\begin{document}
	\maketitle
	\begingroup
	\renewcommand{\thefootnote}{}
	\footnotetext{© 2026 IEEE. Personal use of this material is permitted. Permission from IEEE must be obtained for all other uses, in any current or future media, including reprinting/republishing this material for advertising or promotional purposes, creating new collective works, for resale or redistribution to servers or lists, or reuse of any copyrighted component of this work in other works.}
	\endgroup
	
\begin{abstract}
	This paper proposes RCMAES, a novel variant of the Covariance Matrix Adaptation Evolution Strategy (CMA-ES) for CEC benchmark optimization. RCMAES integrates a dimension-dependent nonlinear population-size reduction strategy with an adaptive restart mechanism within a pure CMA-ES framework. RCMAES is evaluated on three benchmark suites (CEC2017, CEC2020, and CEC2022) and compared with state-of-the-art DE algorithms as well as its closely related counterpart, BIPOP-aCMAES. Experimental results show that RCMAES achieves competitive and robust performance across all benchmarks.
\end{abstract}




\section{Introduction}

The IEEE Congress on Evolutionary Computation (CEC) competitions on bound-constrained single-objective optimization provide a standard testbed for evaluating evolutionary algorithms under fixed function evaluation budgets. In recent editions, Differential Evolution (DE) and its variants have been the most frequently adopted approaches.

The Covariance Matrix Adaptation Evolution Strategy (CMA-ES)~\cite{6790628} follows a distinct optimization paradigm based on learning second-order information via covariance matrix adaptation. Although CMA-ES is a well-established and widely studied algorithm, pure CMA-ES variants are less commonly used as competitive submissions in recent CEC bound-constrained competitions.

Motivated by competition-oriented advances in DE, this work revisits CMA-ES from a CEC competition perspective. Inspired by the Adaptive Restart--Refine Differential Evolution (ARRDE) algorithm~\cite{muzakka2026robustdifferentialevolutionnonlinear}, we incorporate a restart mechanism with nonlinear, dimension-dependent population-size reduction into a pure CMA-ES framework. Unlike IPOP- and BIPOP-aCMA-ES~\cite{1554902, hansen:inria-00382093}, which increase population size after restarts, the proposed strategy favors repeated restarts with the same or reduced population sizes and initializes new search phases away from previously converged regions.

Based on these ideas, we propose \textit{RCMAES} (population-reduction Active CMA-ES with Restart), a CMA-ES variant tailored for bound-constrained optimization in the CEC competition setting. The algorithm is evaluated comprehensively on the CEC2017, CEC2020, and CEC2022 benchmark suites, which together cover a wide range of problem dimensionalities and evaluation budgets. In particular, CEC2017 includes higher-dimensional settings (\(D=10, 30, 50,\) and \(100\)), while CEC2020 and CEC2022 focus on lower to medium dimensions ($D\leq 20$) with larger relative evaluation budgets, with CEC2020 providing the largest evaluation budget among the three suites.

To support reproducibility, all algorithms and CEC benchmark problems used in this work, including RCMAES, are implemented within the Minion framework~\cite{muzakka_2025_14893994} and are publicly available. Experimental results generated in this study can be shared upon reasonable request.

\section{CMA-ES}\label{cmaes}

The Covariance Matrix Adaptation Evolution Strategy (CMA-ES)~\cite{6790628} is a stochastic, derivative-free evolutionary algorithm designed for continuous black-box optimization. CMA-ES performs search by iteratively sampling candidate solutions from a multivariate normal distribution and adapting its parameters to capture the underlying structure of the objective-function landscape. 

\subsection{Sampling and Mean Update}

At generation $t$, CMA-ES samples $\lambda$ offspring according to
\begin{equation}
\mathbf{x}_k^{(t)} \sim \mathcal{N}(\mathbf{m}^{(t)}, \sigma^{(t)^2} \mathbf{C}^{(t)}),
\end{equation}
where $\mathbf{m}^{(t)}$ is the distribution mean, $\sigma^{(t)}$ is the global step size, and $\mathbf{C}^{(t)}$ is the covariance matrix.

After evaluating all offspring, the mean is updated by weighted recombination of the best $\mu$ individuals:
\begin{equation}
\mathbf{m}^{(t+1)} = \sum_{i=1}^{\mu} w_i \mathbf{x}_{i:\lambda}^{(t)},
\end{equation}
where $\mathbf{x}_{i:\lambda}^{(t)}$ denotes the $i$-th best solution and $w_i$ are positive weights satisfying $\sum_i w_i = 1$.

\subsection{Covariance Matrix Adaptation}

The covariance matrix is updated using a combination of rank-one and rank-$\mu$ updates:
\begin{equation}\label{C}
\mathbf{C}^{(t+1)} =
(1 - c_1 - c_\mu)\mathbf{C}^{(t)}
+ c_1 \mathbf{p}_c \mathbf{p}_c^\top
+ c_\mu \sum_{i=1}^{\mu} w_i \mathbf{y}_i \mathbf{y}_i^\top,
\end{equation}
where $\mathbf{p}_c$ is the evolution path for covariance adaptation,
$\mathbf{y}_i = (\mathbf{x}_{i:\lambda}^{(t)} - \mathbf{m}^{(t)}) / \sigma^{(t)}$,
and $c_1$ and $c_\mu$ are learning rates.

This update allows CMA-ES to learn correlations between variables and adapt the search distribution to the local topology of the objective function.

\subsection{Step-Size Adaptation}

CMA-ES employs cumulative step-size adaptation (CSA) to control the global step size:
\begin{equation}
\sigma^{(t+1)} = \sigma^{(t)} \exp\left(
\frac{c_\sigma}{d_\sigma}
\left(
\frac{\|\mathbf{p}_\sigma\|}{E\|\mathcal{N}(\mathbf{0}, \mathbf{I})\|} - 1
\right)
\right),
\end{equation}
where $\mathbf{p}_\sigma$ is the evolution path for step-size control, and $c_\sigma$ and $d_\sigma$ are learning parameters. This mechanism enables self-adaptive control of the exploration--exploitation trade-off without manual tuning.

\subsection{Active Covariance Matrix Adaptation}
Instead of using (\ref{C}), one can also use active covariance adaptation~\cite{10.1145/1830761.1830789}, which extends the standard covariance update by incorporating information from unsuccessful search directions. Negative recombination weights are assigned to poorly performing solutions, yielding the modified update:
\begin{align}
\mathbf{C}^{(t+1)} &=
(1 - c_1 - c_\mu^+ - c_\mu^-)\mathbf{C}^{(t)}
+ c_1 \mathbf{p}_c \mathbf{p}_c^\top \nonumber \\
& + c_\mu^+ \sum_{i=1}^{\mu} w_i^+ \mathbf{y}_i \mathbf{y}_i^\top
- c_\mu^- \sum_{j=\mu+1}^{\lambda} |w_j^-| \mathbf{y}_j \mathbf{y}_j^\top,
\end{align}
where $w_i^+ > 0$ and $w_j^- < 0$ denote positive and negative recombination weights, respectively. Active covariance adaptation accelerates learning by explicitly decreasing variance in unpromising directions and has been shown to improve convergence speed and robustness\cite{10.1145/1830761.1830789}.

\begin{algorithm}[t]
\caption{RCMAES (Restart CMA-ES with Population Reduction)}
\label{alg:rcmaes}
\begin{algorithmic}[1]
\State \textbf{Input:} $f$, $D$, bounds $[L,U]$, budget $N_{\max}$
\State $N_{\text{evals}} \gets 0$; $(x^\star,f^\star) \gets (\emptyset,+\infty)$; $\varepsilon \gets 10^{-12}$
\State Compute $N_0$ by~(\ref{N0}) and exponent $r$ by~(\ref{r-formula})

\While{$N_{\text{evals}} < N_{\max}$} \Comment{restart loop}
    \State Sample mean $\mathbf{m}$ uniformly in $[L,U]$, avoiding previous convergence regions if any
    \State Initialize $\sigma \gets 0.3$, $\mathbf{C} \gets \mathbf{I}$, $\mathbf{p}_c,\mathbf{p}_\sigma \gets \mathbf{0}$

    \While{$N_{\text{evals}} < N_{\max}$} \Comment{one CMA-ES run}
        \State $t \gets N_{\text{evals}}/N_{\max}$
        \State $N_p \gets N_0 - (N_0-D)\left[1-(1-t)^r\right]$
        \State Sample and evaluate $N_p$ offspring using CMA-ES
        \State Apply bound-constraint handling
        \State $N_{\text{evals}} \gets N_{\text{evals}} + N_p$
        \State Update $(x^\star,f^\star)$
        \State Update $\mathbf{m}$, $\sigma$, and $\mathbf{C}$ using active CMA-ES

        \If{$(f_{\max}-f_{\min})/\max(|f_{\text{mean}}|,\varepsilon) \le 10^{-8}$}
            \State Store converged mean
            \State \textbf{break} \Comment{terminate current run and restart}
        \EndIf
    \EndWhile
\EndWhile

\State \textbf{return} $x^\star$
\end{algorithmic}
\end{algorithm}

\section{The Proposed RCMAES Algorithm}\label{rcmaes_sec}
The design of RCMAES is motivated by the trade-off between population size and evaluation budget. While larger populations improve exploration in high-dimensional problems, they quickly exhaust the available budget, requiring population-size reduction. Unlike many DE variants (e.g., LSHADE~\cite{b6900380}) that use linear reduction, RCMAES adopts a nonlinear, dimension-dependent strategy. More aggressive reduction is applied in lower dimensions, where large populations are less critical; however, this increases the risk of premature convergence, which is mitigated by an explicit restart mechanism.

Given a maximum evaluation budget $N_{\max}$ and problem dimensionality $D$, RCMAES initializes the population by sampling individuals from a Gaussian distribution centered at a randomly selected mean within the search bounds. The initial population size $N_0$ is defined as
\begin{align}
	N_{0} & = D \times \max\!\left[2,\; 10 \eta - 20 \right], \label{N0} \\
	\eta  & = \log_{10}(N_{\max}/D),
\end{align}
making it dependent on both the available evaluation budget and the problem dimension. This functional form allows larger initial populations when sufficient evaluations are available, while remaining conservative under tight budgets.

The population is then evolved according to the standard CMA-ES update rules. During optimization, the population size is reduced in a nonlinear, time-dependent manner. Let $N_{\text{evals}}$ denote the number of function evaluations used so far, and define $t = N_{\text{evals}} / N_{\max}$. The population size at time $t$ is given by
\begin{align}
	N_p(t) &=
	N_{0} - \left(N_{0} - D\right)\!\left[1 - \left(1 - t\right)^{r}\right], \label{Np} \\
	r & = 1.7 - 0.01D. \label{r-formula}
\end{align}
This formulation yields more aggressive population reduction for low-dimensional problems and more conservative reduction for higher-dimensional ones. In particular, when $r = 1$, the schedule reduces to a linear scheme.

The coefficients in (\ref{N0}) and (\ref{r-formula}) were obtained through empirical tuning on the CEC2017, CEC2020, and CEC2022 benchmark suites. Notably, the proposed formulation implies that $r \geq 1$ (i.e., faster-than-linear reduction) occurs when $D \leq 70$, whereas for $D > 70$, the reduction becomes slower than linear. For example, at $D = 100$, $r = 0.7$, resulting in a significantly more gradual reduction. 

We observed that this dimension-dependent schedule provides a favorable balance between exploration and exploitation. In lower-dimensional settings, more aggressive reduction leads to more frequent restarts, which enhances the ability to escape local optima. Empirical observations (from an ablation study, omitted due to space constraints) indicate that this behavior is particularly beneficial for complex, multimodal problems in the CEC2020 benchmark suite.

A restart is triggered when population convergence is detected, defined as $
\delta = (f_{\max} - f_{\min})/\max(|f_{\text{mean}}|, \varepsilon)\leq 10^{-8}$,
where $\varepsilon=10^{-12}$ and $f_{\max}$, $f_{\min}$, and $f_{\text{mean}}$ denote the maximum, minimum, and mean objective values in the current population, respectively. Upon restart, the population size continues to follow (\ref{Np}), while the covariance matrix, step size, and evolution paths are reset. The new mean is sampled outside a local exclusion region defined as a hyper-rectangular box centered at the previous converged mean, with side lengths equal to $10\%$ of the variable bounds. This mechanism encourages exploration of previously unexplored regions of the search space.

To handle bound constraints, RCMAES adopts a stochastic repair strategy, as the original CMA-ES is designed for unconstrained optimization and does not provide native support for bound constraints. When a candidate violates the search bounds, the infeasible coordinate is re-sampled uniformly within a truncated interval adjacent to the violated boundary. The width of this interval is limited by the violation magnitude, ensuring small corrective steps while preserving diversity near the boundary.

The algorithm terminates when the evaluation budget $N_{\max}$ is exhausted. Pseudocode for RCMAES is provided in Algorithm~\ref{alg:rcmaes}.

\section{Numerical Experiments and Discussions}

\subsection{Experimental Setup}
The proposed RCMAES is evaluated on three benchmark suites: CEC2017~\cite{cec2017}, CEC2020~\cite{cec2020}, and CEC2022~\cite{cec2022}. For all experiments, the maximum number of function evaluations ($N_{\max}$) follows the official competition specifications. Each algorithm is executed for 51 independent runs, using the run index as the random seed to ensure reproducibility. All experiments are conducted within the Minion framework~\cite{muzakka_2025_14893994}, implemented in C++ and compiled with MSVC on a Windows~11 workstation. All function evaluations are performed using standard double-precision floating-point arithmetic. Reported results are presented with standard numerical formatting for readability, which may affect the ranking in cases where performance differences are extremely small.

RCMAES is compared against representative state-of-the-art DE and CMA-ES algorithms, including BIPOP-aCMA-ES~\cite{hansen:inria-00382093}, ARRDE~\cite{muzakka2026robustdifferentialevolutionnonlinear}, jSO~\cite{7969456}, j2020~\cite{9185551}, LSRTDE~\cite{10611907}, and NL-SHADE-RSP~\cite{9504959}. All algorithms use the parameter settings recommended in their original publications.

\subsection{Scoring Methodology}

Algorithm performance is evaluated using both \emph{rank-based} and \emph{accuracy-based} metrics, following standard CEC evaluation practices. Let $k$ denote the algorithm index, $j$ the benchmark problem, $i$ the independent run, $D$ the problem dimension, $N_p$ the number of benchmark problems, and $N_{\text{runs}}$ the number of independent runs.

For rank-based assessment, Friedman ranking~\cite{Friedman01121937} is employed. For each problem and run, algorithms are ranked according to their final objective values, with smaller values indicating better performance; ties are assigned the average rank. Let $r_{i,j,k}$ denote the rank of algorithm $k$ on problem $j$ in run $i$. The overall Friedman score is defined as
\[
R_k
= \frac{1}{N_p\,N_{\text{runs}}}
\sum_{j=1}^{N_p} \sum_{i=1}^{N_{\text{runs}}} r_{i,j,k}.
\]
Smaller values of $R_k$ indicate superior overall performance.

To assess whether one algorithm outperforms another in terms of win, tie, or loss outcomes, pairwise win/tie/loss (W/T/L) statistics are employed. For each pair of algorithms, results from $N_{\text{runs}}$ independent runs are compared using the Mann--Whitney~U test~\cite{10.1214/aoms/1177730491} at a significance level of $\alpha=0.05$. If no statistically significant difference is detected, the comparison is recorded as a tie. Otherwise, the direction of preference is determined according to the rank-sum statistic, yielding either a win or a loss.

Accuracy-based performance is measured using the relative error. For algorithm $k$ on problem $j$, the relative error is defined as
\[
\epsilon_{k,j}
= \frac{1}{N_{\text{runs}}}
\sum_{i=1}^{N_{\text{runs}}}
\frac{f_j(x_{i,k,j}) - f_j(x_j^*)}{f_j(x_j^*)},
\]
where $x_{i,k,j}$ is the solution obtained by algorithm $k$ in run $i$, $f_j(\cdot)$ is the objective function, and $x_j^*$ is the known global optimum. Note that for the considered benchmark suites (CEC2017, CEC2020, and CEC2022), all function values are strictly positive, ensuring that the relative error is well-defined.

To limit the influence of outlier problems, the relative error is mapped to a bounded interval via
\[
\mathcal{E}_{k,j} = \frac{\epsilon_{k,j}}{1 + \epsilon_{k,j}},
\]
which restricts values to $[0,1)$. This transformation retains sensitivity for small errors while preventing large errors from dominating the aggregated score. The overall accuracy-based score $\mathcal{E}_k$ of algorithm $k$ is obtained by averaging $\mathcal{E}_{k,j}$ over all benchmark problems for the given dimension.

\subsection{Results}

	\begin{table*}[ht]
	\centering
	\footnotesize
	\setlength{\tabcolsep}{3pt} 
	\caption{CEC2017 results at $D=10, 30, 50, 100$ showing accuracy \(\mathcal{E}\), Friedman rank \(R\), and win/tie/loss (W/T/L) statistics of RCMAES against competing algorithms. Values in parentheses denote ranks; boldface indicates the best performance.}
		\label{2017}
	\begin{tabular}{l|ccc|ccc|ccc|ccc}
		\hline
		Algorithm & \multicolumn{3}{|c|}{10D} & \multicolumn{3}{|c|}{30D} & \multicolumn{3}{|c|}{50D} & \multicolumn{3}{|c}{100D} \\
		& $\mathcal{E}$ & $R$ & W/T/L & $\mathcal{E}$ & $R$ & W/T/L & $\mathcal{E}$ & $R$ & W/T/L & $\mathcal{E}$ & $R$ & W/T/L \\
\hline
ARRDE & \textbf{0.031 (1)} & \textbf{3.349 (1)} & 5/11/13 & 0.094 (4) & 3.445 (3) & 13/13/3 & 0.160 (4) & 3.402 (3) & 20/8/1 & 0.252 (3) & 3.438 (3) & 22/1/6 \\
jSO & 0.036 (4) & 3.620 (2) & 10/9/10 & 0.092 (3) & 3.452 (4) & 19/8/2 & 0.157 (3) & 3.419 (4) & 18/9/2 & 0.260 (4) & 3.523 (4) & 21/2/6 \\
j2020 & 0.037 (5) & 4.577 (6) & 13/10/6 & 0.200 (7) & 6.016 (7) & 25/4/0 & 0.403 (7) & 6.139 (7) & 26/2/1 & 0.596 (7) & 6.318 (7) & 25/3/1 \\
NL-SHADE-RSP & 0.035 (2) & 3.894 (5) & 11/6/12 & 0.168 (6) & 5.144 (6) & 23/3/3 & 0.362 (6) & 5.886 (6) & 26/2/1 & 0.546 (6) & 6.035 (6) & 27/1/1 \\
LSRTDE & 0.063 (7) & 3.716 (3) & 10/9/10 & \textbf{0.068 (1)} & \textbf{2.396 (1)} & 12/11/6 & \textbf{0.101 (1)} & \textbf{2.083 (1)} & 14/9/6 & 0.199 (2) & \textbf{2.125 (1)} & 16/4/9 \\
RCMAES & 0.035 (3) & 3.731 (4) & -- & 0.071 (2) & 2.436 (2) & -- & 0.129 (2) & 2.139 (2) & -- & \textbf{0.190 (1)} & 2.261 (2) & -- \\
BIPOP-aCMAES & 0.051 (6) & 5.114 (7) & 21/8/0 & 0.164 (5) & 5.111 (5) & 25/3/1 & 0.262 (5) & 4.931 (5) & 24/3/2 & 0.354 (5) & 4.300 (5) & 22/2/5 \\
\hline
	\end{tabular}
\end{table*}

\paragraph{CEC2017}
Table~\ref{2017} summarizes the results on the CEC2017 benchmark suite for \(D=10\), \(30\), \(50\), and \(100\) using W/T/L statistics, Friedman rank \(R\), and the accuracy-based metric \(\mathcal{E}\). Since these metrics capture different aspects of performance, they should be interpreted jointly.

Across all dimensions, RCMAES exhibits a strong pairwise W/T/L profile. At \(D=10\), RCMAES performs competitively against all algorithms, with the exception of ARRDE and NL-SHADE-RSP. At higher dimensions (\(D=30\), \(50\), and \(100\)), RCMAES demonstrates pairwise superiority, winning substantially more often against most competing algorithms.

Notably, RCMAES outperforms LSRTDE in direct pairwise comparisons, despite LSRTDE being a particularly strong baseline. In its original publication\cite{10611907}, LSRTDE was shown to exhibit decisive and systematic dominance over the CEC2017 winning algorithms, such as jSO, establishing a clear performance gap rather than marginal improvements. The W/T/L statistics reported in Table~\ref{2017} are consistent with this characterization, as LSRTDE continues to significantly outperform these earlier winners. Against this strong reference, RCMAES nonetheless achieves more frequent wins in direct comparisons.

In terms of Friedman ranking, RCMAES ranks second at \(D=30\), \(50\), and \(100\), and third at \(D=10\), while LSRTDE attains the top rank for \(D \geq 30\). The apparent discrepancy between the strong W/T/L results and the Friedman ranks can be attributed to the fundamentally different nature of these measures. Friedman ranking aggregates relative ordering across all algorithms, such that poor performance on a small number of functions can disproportionately affect the overall rank. In the case of RCMAES, this effect is primarily due to function F29, on which RCMAES performs relatively poorly. In contrast, W/T/L analysis reflects pairwise statistical significance. Indeed, when restricting the comparison to RCMAES and LSRTDE only, or when excluding F29, we verified that RCMAES achieves a better Friedman rank than LSRTDE. Overall, these results indicate that the performance of RCMAES and LSRTDE is very close, with differences largely attributable to isolated functions rather than systematic superiority.

The accuracy-based metric \(\mathcal{E}\) provides complementary insight into solution quality. LSRTDE achieves the lowest error values at \(D=30\) and \(D=50\), with RCMAES ranking closely behind and attaining the lowest error at \(D=100\). The accuracy differences between these two algorithms are small, while both clearly outperform the remaining methods. At \(D=10\), ARRDE achieves the best accuracy, with RCMAES yielding comparable error values, whereas LSRTDE ranks last under this metric.

\begin{center}
	\begin{table*}[ht]
		\centering
		\footnotesize
		\setlength{\tabcolsep}{3pt} 
		\caption{CEC2020 results at $D=5, 10, 15, 20$ showing accuracy \(\mathcal{E}\), Friedman rank \(R\), and win/tie/loss (W/T/L) statistics of RCMAES against competing algorithms. Values in parentheses denote ranks; boldface indicates the best performance.}
		\label{2020}
		\begin{tabular}{l|ccc|ccc|ccc|ccc}
		\hline
		Algorithm & \multicolumn{3}{|c|}{5D} & \multicolumn{3}{|c|}{10D} & \multicolumn{3}{|c|}{15D} & \multicolumn{3}{|c}{20D} \\
		& $\mathcal{E}$ & $R$ & W/T/L & $\mathcal{E}$ & $R$ & W/T/L & $\mathcal{E}$ & $R$ & W/T/L & $\mathcal{E}$ & $R$ & W/T/L \\
	\hline
	ARRDE & \textbf{0.008 (1)} & 3.372 (2) & 1/3/6 & \textbf{0.009 (1)} & 3.425 (3) & 2/3/5 & \textbf{0.018 (1)} & 3.463 (2) & 3/3/4 & \textbf{0.023 (1)} & \textbf{3.378 (1)} & 4/1/5 \\
	jSO & 0.018 (4) & 4.097 (4) & 1/4/5 & 0.029 (5) & 4.397 (6) & 4/2/4 & 0.039 (7) & 4.934 (6) & 6/2/2 & 0.036 (6) & 4.961 (6) & 6/2/2 \\
	j2020 & 0.010 (2) & 3.837 (3) & 2/2/6 & 0.011 (2) & 3.374 (2) & 2/2/6 & 0.027 (4) & 3.885 (5) & 4/3/3 & 0.039 (7) & 3.786 (5) & 3/3/4 \\
	NL-SHADE-RSP & 0.010 (3) & \textbf{2.875 (1)} & 0/2/8 & 0.018 (3) & \textbf{3.154 (1)} & 1/4/5 & 0.022 (2) & \textbf{3.103 (1)} & 1/3/6 & 0.032 (3) & 3.393 (2) & 4/2/4 \\
	LSRTDE & 0.018 (5) & 4.560 (6) & 2/7/1 & 0.032 (7) & 5.579 (7) & 8/2/0 & 0.038 (6) & 5.458 (7) & 8/2/0 & 0.035 (5) & 5.481 (7) & 9/1/0 \\
	RCMAES & 0.019 (6) & 4.519 (5) & -- & 0.023 (4) & 4.111 (5) & -- & 0.024 (3) & 3.680 (4) & -- & 0.024 (2) & 3.527 (4) & -- \\
	BIPOP-aCMAES & 0.020 (7) & 4.741 (7) & 3/6/1 & 0.030 (6) & 3.961 (4) & 3/2/5 & 0.028 (5) & 3.476 (3) & 2/3/5 & 0.032 (4) & 3.473 (3) & 2/4/4 \\
	\hline
		\end{tabular}
	\end{table*}
\end{center}

\paragraph{CEC2020}
Table~\ref{2020} summarizes the results on the CEC2020 benchmark suite for \(D=5\), \(10\), \(15\), and \(20\). Compared to its performance on CEC2017, RCMAES is less competitive on this benchmark. However, all three performance indicators—Friedman rank \(R\), accuracy \(\mathcal{E}\), and W/T/L statistics—exhibit consistent improvement as the problem dimension increases. Notably, at \(D=15\) and \(D=20\), RCMAES performs competitively relative to j2020, despite j2020 being specifically designed for the CEC2020 benchmark.

A marked contrast with the CEC2017 results is observed when examining overall algorithm behavior. LSRTDE, which achieved strong rankings on CEC2017, ranks last or near last across all dimensions on CEC2020. In contrast, algorithms that performed less favorably on CEC2017, such as j2020 and NL-SHADE-RSP, achieve substantially improved rankings on CEC2020, with NL-SHADE-RSP consistently ranking among the top methods. These observations are consistent with the findings reported in~\cite{PIOTROWSKI2023101378, PIOTROWSKI2025101807}.

Although RCMAES experiences a degradation in overall ranking on CEC2020, this degradation is relatively moderate, indicating better robustness compared to the competing algorithms.

\begin{center}
	\begin{table*}[ht]
		\centering
		\footnotesize 
		\caption{CEC2022 results for $D=10$ and $D=20$, reporting accuracy \(\mathcal{E}\), Friedman rank \(R\), and win/tie/loss (W/T/L) statistics of RCMAES against competing algorithms. Values in parentheses denote ranks; boldface indicates the best performance.
		}
			\label{2022}
		\begin{tabular}{l|ccc|ccc}
		\hline
		Algorithm & \multicolumn{3}{|c|}{10D} & \multicolumn{3}{|c}{20D} \\
		& $\mathcal{E}$ & $R$ & W/T/L & $\mathcal{E}$ & $R$ & W/T/L \\
		\hline
		ARRDE & 0.014 (3) & \textbf{3.516 (1)} & 2/4/6 & 0.017 (2) & 3.320 (2) & 3/6/3 \\
		jSO & 0.016 (5) & 3.992 (4) & 2/6/4 & 0.035 (6) & 4.227 (5) & 5/6/1 \\
		j2020 & \textbf{0.013 (1)} & 3.660 (2) & 2/4/6 & 0.026 (4) & 4.696 (6) & 7/4/1 \\
		NL-SHADE-RSP & 0.014 (2) & 3.790 (3) & 2/6/4 & 0.038 (7) & 4.876 (7) & 7/4/1 \\
		LSRTDE & 0.016 (6) & 4.233 (6) & 4/6/2 & 0.034 (5) & 4.069 (4) & 5/7/0 \\
		RCMAES & 0.016 (4) & 4.029 (5) & -- & \textbf{0.016 (1)} & \textbf{3.170 (1)} & -- \\
		BIPOP-aCMAES & 0.021 (7) & 4.780 (7) & 4/7/1 & 0.023 (3) & 3.642 (3) & 5/3/4 \\
		\hline
		\end{tabular}
	\end{table*}
\end{center}

\paragraph{CEC2022}
Table~\ref{2022} summarizes the results on the CEC2022 benchmark suite for \(D=5\), \(10\), \(15\), and \(20\). RCMAES achieves the best performance at \(D=20\), ranking first according to both the accuracy metric \(\mathcal{E}\) and the Friedman rank \(R\). At \(D=10\), however, RCMAES ranks fourth in terms of \(\mathcal{E}\) and fifth according to \(R\).

\subsection{CEC2026}
\begin{figure*}[t]
	\centering
	\includegraphics[width=\linewidth]{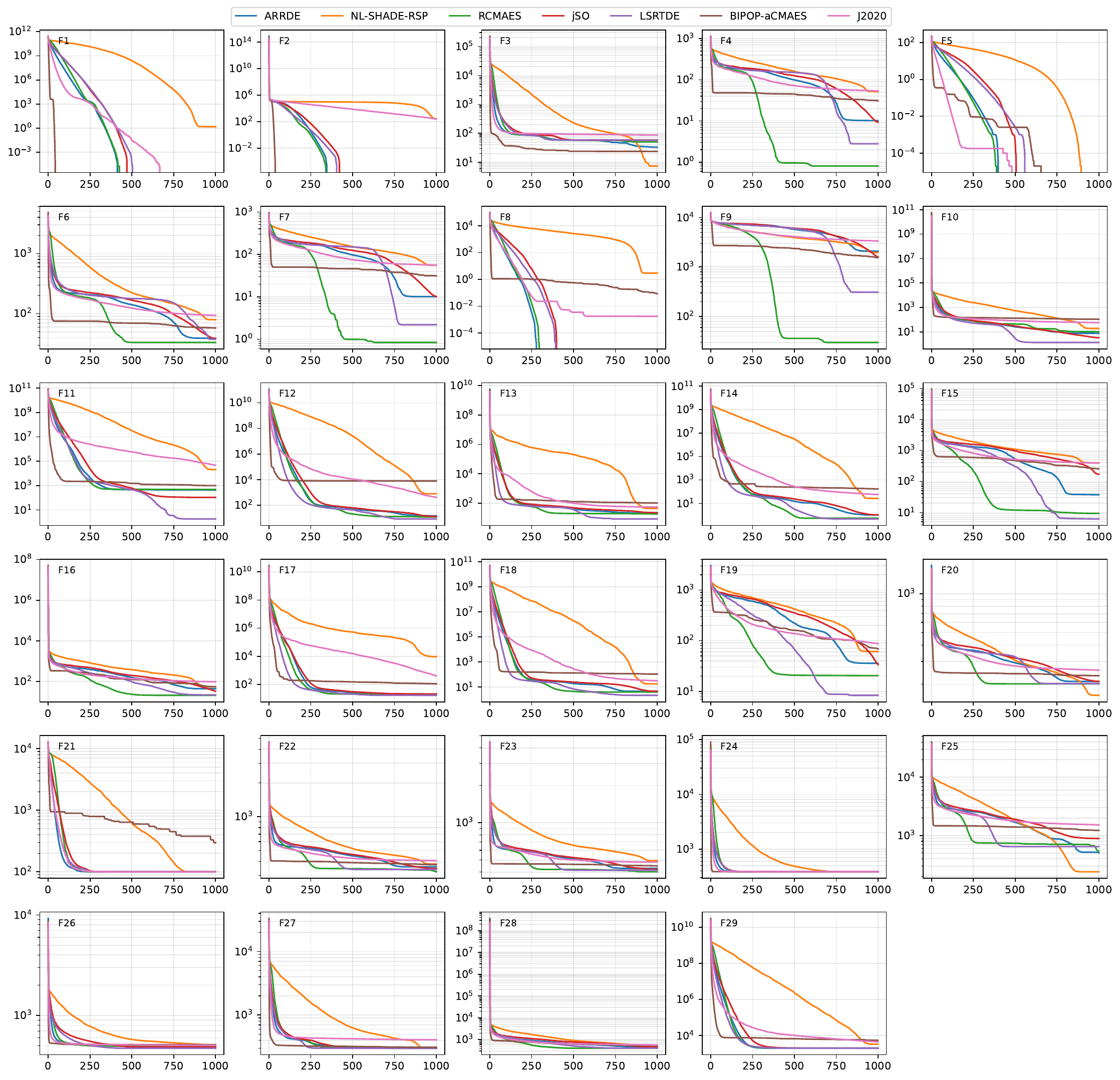}
	\caption{Average convergence curves on the CEC2017 benchmark at \(D=30\). The plots show the mean error $f-f^*$ over all runs as a function of the normalized number of function evaluations \(N_{\mathrm{evals}}/D\).}
	\label{converg}
\end{figure*}

The IEEE CEC2026 competition on single-objective bound-constrained optimization adopts the CEC2017 benchmark suite at \(D=30\). The competition emphasizes both fast convergence toward the global optimum and high final solution accuracy. While the final accuracy has been reported in Table~\ref{2017}, this subsection focuses on convergence behavior, illustrated in Fig.~\ref{converg}, which shows the average error $f-f^*$ across independent runs as a function of the normalized evaluation budget \(N_{\mathrm{evals}}/D\).

From the convergence curves, RCMAES demonstrates strong performance on basic functions, particularly F4, F6, F7, and F9, where it converges rapidly toward high-quality solutions. In contrast, LSRTDE shows clear strengths on hybrid functions, such as F11, F13, and F15.

Algorithms that perform well on CEC2020, such as NL-SHADE-RSP and j2020, exhibit comparatively slow convergence on CEC2017 at \(D=30\). Meanwhile, BIPOP-aCMAES converges rapidly in the early stages but frequently stagnates at local optima, resulting in final solutions that remain far from the global optimum.

For composition functions, most algorithms display similar convergence trends. RCMAES, however, occasionally achieves faster convergence than competing methods, as observed on functions F22 and F23.

Following the CEC2026 competition guidelines, time complexity is evaluated using two measures: \(T_1\), the average time required to evaluate each of the 29 benchmark functions in 30D for 10\,000 evaluations, and \(T_2\), the average time required for the algorithm to run with 10\,000 function evaluations. For RCMAES, the measured values are \(T_1 = 0.0453\), \(T_2 = 0.0802\), resulting in a normalized computational overhead of \((T_2 - T_1)/T_1 = 0.7705\).

\section{Conclusion}
This paper introduced RCMAES, a CMA-ES variant that integrates a dimension-dependent nonlinear population-size reduction strategy with an adaptive restart mechanism. Experimental results on the CEC2017, CEC2020, and CEC2022 benchmark suites demonstrate that RCMAES achieves competitive and robust performance compared with state-of-the-art algorithms. Notably, RCMAES exhibits stable behavior across benchmark suites with differing characteristics.

\section*{Acknowledgments}

The authors acknowledge financial support from the German Federal Ministry of Education and Research (BMBF) under Grant No.~13XP0445 (For-Analytik). K.F.M., S.M., and M.F. are supported by this grant. 

The authors acknowledge the use of artificial intelligence tools (ChatGPT, OpenAI) for language refinement and clarity improvement during manuscript preparation. All technical content, experiments, and conclusions were developed and verified by the authors.

\bibliographystyle{unsrt}
\bibliography{ref}


\end{document}